\documentclass{article}

\usepackage[final]{neurips_2023}

\usepackage[utf8]{inputenc} 
\usepackage[T1]{fontenc}    
\usepackage{hyperref}       
\usepackage{url}            
\usepackage{booktabs}       
\usepackage{amsfonts}       
\usepackage{nicefrac}       
\usepackage{microtype}      
\usepackage{xcolor}         
\usepackage{amssymb}
\usepackage{graphicx}
\usepackage{amsmath}
\usepackage{cleveref}
\usepackage{listings}
\usepackage{latexsym}
\usepackage{tabularx}
\usepackage{booktabs}
\usepackage{color, soul}
\usepackage{algorithm}
\usepackage{algorithmic}
\usepackage{xspace}
\usepackage{wrapfig,lipsum}
\usepackage{graphicx}
\usepackage{subcaption}

\newcommand{\CHATGPT}{ChatGPT\xspace}
\newcommand{\GPT}{GPT4\xspace}
\newcommand{\GPTJDCnR}{GPTJ-D-CnR\xspace}
\newcommand{\GPTNeoXDCnR}{GPT-NeoX-D-CnR\xspace}
\newcommand{\FalconDCnR}{Falcon-D-CnR\xspace}
\newcommand{\FalconICnR}{Falcon-I-CnR\xspace}

\definecolor{codegreen}{rgb}{0,0.6,0}
\definecolor{codegray}{rgb}{0.5,0.5,0.5}
\definecolor{codepurple}{rgb}{0.58,0,0.82}
\definecolor{backcolour}{rgb}{0.95,0.95,0.92}
\definecolor{darkred}{rgb}{0.6,0.0,0.0}
\definecolor{darkgreen}{rgb}{0,0.50,0}
\definecolor{lightblue}{rgb}{0.0,0.42,0.91}
\definecolor{orange}{rgb}{0.99,0.48,0.13}
\definecolor{grass}{rgb}{0.18,0.80,0.18}
\definecolor{pink}{rgb}{0.97,0.15,0.45}

\lstdefinestyle{mystyle}{
  frame=single,
  basicstyle=\ttfamily\footnotesize,
  backgroundcolor=\color{backcolour}, commentstyle=\color{codegreen},
  commentstyle=\color{darkgreen}\slshape,
  keywordstyle=\color{blue},
  stringstyle=\color{darkred},
  numberstyle=\tiny\color{codegray},
  emphstyle=\color{pink}\underbar,
  morekeywords={Verify, Question},
  escapeinside={(*@}{@*)},
  breakatwhitespace=false,         
  breaklines=false,                 
  captionpos=b,                    
  keepspaces=true,                    
  numbersep=5pt,                  
  showspaces=false,                
  showstringspaces=false,
  showtabs=false,                  
  tabsize=2
}

\title{Data-Efficient Alignment of Large Language Models with Human Feedback Through Natural Language}

\author{
    Di Jin,
    Shikib Mehri,
    Devamanyu Hazarika,
    Aishwarya Padmakumar \\
    \textbf{Sungjin Lee, Yang Liu, Mahdi Namazifar}
    \\ \\
    Amazon\\
    \texttt{\{djinamzn,asmehri,dvhaz,padmakua,sungjinl,yangliud,mahdinam\}@amazon.com}
}

\begin{document}

\maketitle

\begin{abstract}
  Learning from human feedback is a prominent technique to align the output of large language models (LLMs) with human expectations. Reinforcement learning from human feedback (RLHF) leverages human preference signals that are in the form of ranking of response pairs to perform this alignment. However, human preference on LLM outputs can come in much richer forms including natural language, which may provide detailed feedback on strengths and weaknesses of a given response. 
In this work we investigate data efficiency of modeling human feedback that is in natural language. 
Specifically, we fine-tune an open-source LLM, e.g., Falcon-40B-Instruct, on a relatively small amount (1000 records or even less) of human feedback in natural language in the form of critiques and revisions of responses. We show that this model is able to improve the quality of responses from even some of the strongest LLMs such as ChatGPT, BARD, and Vicuna, through critique and revision of those responses. For instance, through one iteration of revision of ChatGPT responses, the revised responses have 56.6\% win rate over the original ones, and this win rate can be further improved to 65.9\% after applying the revision for five iterations.
\end{abstract}

\section{Introduction}
Recent developments in Large Language Models (LLMs) have deeply impacted the course of research and practice in Natural Language Processing (NLP), conversational AI, and beyond, since the introduction of ChatGPT. Currently there are three general steps in building such LLM-based conversational systems, which are pre-training, supervised fine-tuning, and further alignment of model responses with human preferences. The last step is often done through Reinforcement Learning from Human Feedback (RLHF)~\cite{NEURIPS2022_b1efde53}. In this approach, first a reward model is trained using human annotations of the preferred responses between pairs of responses. The reward model is then used to further fine-tune the LLM through RL, e.g., using online policy optimization, PPO~\cite{schulman2017proximal}. 


Aligning outputs of an LLM with human preferences using RL and based on a reward model has its own challenges. RLHF is notoriously unstable during training and requires much more computational resources to host both policy and base models, which imposes great challenges to the infrastructure when the model size exceeds 50B parameters~\cite{lu2022quark}. Additionally, in RLHF human preference is conveyed to the model through a single reward value. However human preferences could be much more elaborate, specific, and detailed. For example, a response might be preferred because it is concise and to the point, or covers a certain key aspect in more details and leaves other aspects at high level. To enable LLMs to learn from fine-grained human feedback beyond a singular value, recent work such as Chain of Hindsight~\cite{liu2023languages}, looked into how to teach the model human preferences without a reward model by showing the model two responses and indicating that one is preferred over the other one based on certain reasons in natural language. 

In this work we teach the model human preferences through examples and in natural language. Specifically, given a prompt and a response, we teach the model to (1) generate a critique of this response and outline the positive and negative aspects of the response, and (2) revise the initial response according to the critique. We refer to this approach as \textbf{Critique and Revise}, or \textbf{CnR} for short.
Through human annotation, given a prompt and a response we collect a critique of the response, as well as a revised response based on the critique. 
With a relatively small set of such annotated samples, e.g., fewer than 1000 samples, CnR is able to teach an open-source LLM like Falcon-40B-Instruct to revise and improve even ChatGPT responses.
Through one iteration of ChatGPT response revision, the revised responses have 56.6\% win rate over the original ones, and this win rate can be further improved to 65.9\% after applying five iterations of revisions.
Such great data efficiency implies that it is easier for models to learn the alignment to human feedback via detailed critiques in natural language compared to pairwise comparison feedback that simply tells which of two responses is preferred.

Furthermore, through our collected diagnostic test set we validate that the CnR model is able to generate critiques that are meaningful and specific to a given response and can address those errors in the response.   We also show that this model can follow a given critique and revise the responses accordingly. Moreover, through ablation studies, we show that the revision quality can be improved by using larger base pre-trained models and higher quality instruction-tuning data. We also study the impact of the amount of CnR data on the performance of the CnR model.



\begin{figure*}[ht!]
\centering
\includegraphics[width=0.9\textwidth]{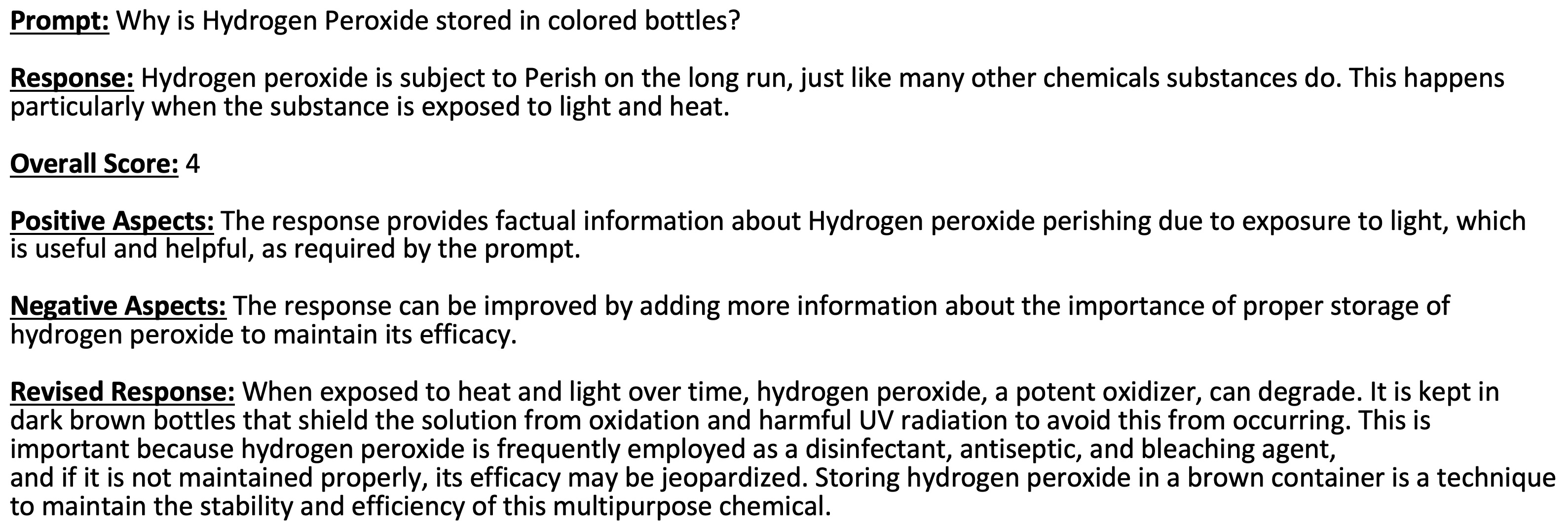}
\caption{Example of critique and revision annotation}
\label{fig:cnr-ann-ex}
\end{figure*}

\section{Related Work}
\label{sec:related-works}

\paragraph{Alignment via Preference Modeling.}

To align LLM outputs further with human preferences or values, a popular approach is to learn and utilize preference models that can mimic human preference. Leveraging these models, LLMs can be trained to generate responses that ensure high preference scores. These approaches require training reward models and reinforcement learning strategies, such as PPO~\cite{NEURIPS2022_b1efde53,glaese2022improving}. 

To alleviate the dependence on complex RL training, recent work attempted to bypass RL via approaches such as incorporating implicit RLHF loss, e.g., DPO~\cite{rafailov2023direct} or introducing ranking-based loss (e.g., Slic~\cite{zhao2022calibrating}, PPO~\cite{song2023preference}, Stable Alignment~\cite{liu2023training}, RRHF~\cite{yuan2023rrhf}).
 
\paragraph{Alignment via Natural Language Feedback.}

While reward-based human preference modeling is appealing, it has the natural crutch in requiring a reward model in the first place, which itself may be noisy and inaccurate. An alternate approach for alignment is to utilize the primary LLM itself for this purpose. Particularly, revision-based approaches are predominant in such setups.

The literature of revising model generations contains a large body of work, such as PEER~\cite{schick2022peer}, which built on the traditional problem of editing text~\cite{dwivedi2022editeval}. However, earlier work studied this problem in domain-specific settings, including for Wikipedia document, code revisions \cite{reid2022learning} or specific tasks such as grammatical error correction, sentence simplification \cite{malmi2022text}. These often require custom feedback or revision generators. 

Recent work, including Self-Refine~\cite{madaan2023self}, Constitutional AI~\cite{bai2022constitutional}, etc., started utilizing the instruction-following capabilities of LLMs to model human feedback in natural language \cite{bai2022constitutional}. Via prompting, these approaches are capable of generating critiques that include multi-faceted attributes (e.g., making responses helpful, harmless,  as well as faithful and honest) and performing iterative revisions to maximally improve model generations. By using the same LLM for both critique and revisions, LLMs can essentially \textit{self-train} to improve themselves. The overall form of revising generations provides an interpretable framework where the revisions can be observed explicitly in natural language. For an extensive review of recent literature on critique and revise frameworks, we refer the readers to~\cite{madaan2023self} and~\cite{pan2023automatically}.

\paragraph{Is Critique Required in CnR?} Feedback or critique generation has multiple use-cases. For example, it can help annotators to identify issues in model generations and assist them in evaluating responses (for preference data collection)~\cite{saunders2022self} or in revising model responses with assistance from the critique. Critiques can also be used for identifying annotation issues, especially with adversarial human annotators. Moreover, recent work such as ~\cite{bai2022constitutional} demonstrated that critiques do provide improved revisions, thus justifying their generation. Furthermore, revision via critique can better evaluate whether the revision improves issues highlighted in the critique. For example, \cite{campos2022training} used similarity as a heuristic to select best revisions with respect to the critique.

\paragraph{Multi-Model Feedback Providers.} A recent line of work utilized multiple LLMs as feedback generators and revisers~\cite{liu2023training}. For example, \cite{fu2023improving} used two LLMs to negotiate with each other via self-play, where a third LLM acted as a critic to advise one of the negotiators. In contrast to these approaches, our work considers the same LLM for both critique and revision~\cite{madaan2023self}.

\paragraph{General Purpose Feedback in Natural Language.}

Most of the recent work that modeled human feedback in natural language studied the problem in either specific applications, like summarization~\cite{saunders2022self}, or 
a combination of them, e.g., code optimization, acronym generation, dialog response generation, math reasoning, etc. in Self-Refine~\cite{madaan2023self}. While the latter has much more coverage of tasks than the former ones, our focus is to study CnR in general purpose human demonstrations. 

\paragraph{Training vs. Few-Shot for Natural Language Feedback Setup.}

Several prior studies have noted that zero- or few-shot critique or revision generation requires large powerful LLMs \cite{saunders2022self, campos2022training, madaan2023self}. However, large LLMs are usually not open-sourced and come with restrictive licenses, prohibiting use-cases that demand privacy needs, e.g., legal and healthcare use cases. Given this issue, our work studies the paradigm of CnR in smaller open-sourced models. 

The recently proposed AlMoST~\cite{kim2023aligning} shares a similar motivation as ours, in that it also attempts to perform alignment without the need for a large scale teacher model. For this, they utilize multiple small scale models and bake in a relative preference of responses from a relatively larger  scale model (e.g., 30B) with few-shot demonstrations to be preferred over a 1B model with zero-shot prompting. However, AlMoST utilizes SFT and RL for alignment, whereas we study the critique and revision framework. \cite{vernikos2023small} also demonstrated that smaller models can act as strong refiners when trained aptly.


\section{CnR Data} \label{sec:data}

In this section we discuss how we collected the data for CnR. For a given prompt and a response, the human annotation task is to provide a critique of the response and a revised response according to the critique. We start with a set of 1,000 human demonstrations that are collected inspired by the taxonomy introduced in InstructGPT \cite{NEURIPS2022_b1efde53} for summarization, generation, question answering, extraction, brainstorming, and rewrite.  Each of these demonstrations includes a prompt and a response, both written by a human annotator. We use 3 different LLMs that are based on FLAN-T5-XXL, GPT-J, and AlexaTM 20B \cite{chung2022scaling, fitzgerald2022alexa} and are fine-tuned on the Anthropic Helpful dataset~\cite{bai2022training}\footnote{We used base, online, and rejection-sampled version in this link: https://github.com/anthropics/hh-rlhf} to generate 3 additional responses for each prompt. Finally, we randomly select one from the 4 responses for each of the 1000 prompts. The resulting dataset is 1000 prompts and responses, where we have a mixed quality of responses. Approximately a quarter of the prompts  have responses written by humans that we expect are of high quality, and for the rest of the prompts, the responses are generated from relatively weaker models trained on subpar data, and are of lower quality. 

We collected critique and revision for this set of prompts and responses. As illustrated by Figure~\ref{fig:cnr-ann-ex}, for each response to a prompt, annotators are asked to give a score from 1 to 5 for the overall quality of the response. They are also asked to specify positive aspects of the response, as well as its negative aspects. In providing a critique for a response, they are asked to refrain from providing generic statements such as ``the overall quality is high'' or ``the response is flawed''. Instead they are instructed to point out specific sentences or phrases in the response that are flawed and specify the reason why those statements are problematic. Moreover, the annotators are instructed not to use first person language such as ``in my opinion'' or ``I feel like'' in their critique of the response. If a response is already good, annotators can indicate that nothing needs to be improved in the ``negative aspects''.

Finally the annotators are asked to provide a revised response for the prompt according to the positive and negative aspects that they mention in the critique.  The revised responses need to be very high quality. 
Moreover, annotators were instructed to ensure that critiques and revisions are fully aligned. In other words, if a negative or a positive point is mentioned in the critique, it should be reflected in the revised response. Also no additional changes should be introduced in the revised response that are not mentioned in the critique of the initial response. All annotated records were reviewed by a second annotator and changes were made when necessary in this review process.

Figure \ref{fig:cnr-ann-ex} shows a CnR annotated example, consisting of a prompt, the initial response, the critique of the initial response (that includes an overall score, a set of positive aspects and a set of negative aspects), and a revised response. 


\section{CnR Model}
\label{sec:model}

\begin{figure*}[ht!]
\centering
\includegraphics[width=1\textwidth]{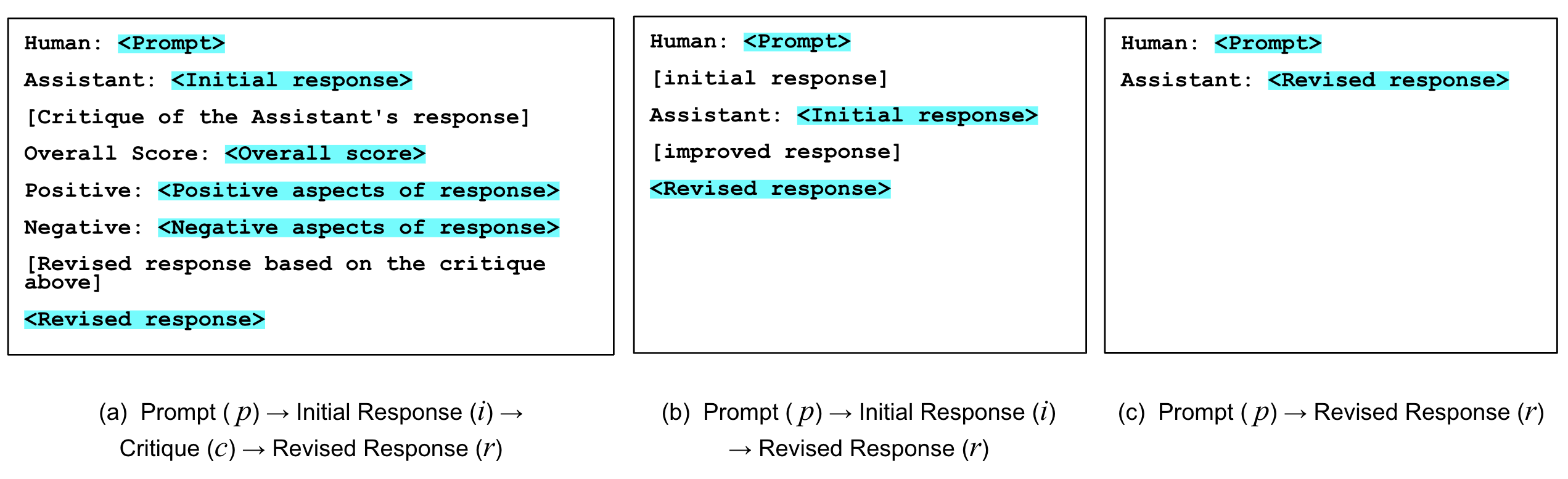}
\caption{Different CnR data preparation settings}
\label{fig:cnr-rec}
\end{figure*}

Training of the CnR model is done in the same way as Supervised Fine-tuning (SFT) of LLMs on human demonstrations \cite{NEURIPS2022_b1efde53}. The critique and revision annotated data is used to create CnR training records. 
In this work we experiment with different ways of constructing these records for model training and inference. Each annotated sample includes a prompt, an initial response, a critique, and a revised response, which are denoted as $p$, $i$, $c$, and $r$, respectively. As illustrated by Figure \ref{fig:cnr-rec}, we consider three formulations to construct the training records: (1)  using all annotations $p \rightarrow i \rightarrow c \rightarrow r$, (2) only using the initial response and the revised responses, without the critiques $p \rightarrow i \rightarrow r$, (3) only using the revised responses $p \rightarrow r$. 
We compare which setting achieves better performance.

For training, an already supervised fine-tuned model is continually fine-tuned on the CnR training records. During this training no masking over the training records is applied while calculating the loss. One benefit of this is that the final CnR model can be used with or without an initial response or a critique. For instance, in this setting one can prompt the final CnR model to generate an initial response, a critique of this initial response (an overall score, positive aspects, and negative aspects), and then a revised response based on the critique. Alternatively, this model can be used to critique and revise a given response to a prompt; or given a prompt, a response, and a critique of the response, the model can be used to generate a revised response.

In this work we run CnR fine-tuning on a number of different open source models including GPT-J~\cite{gpt-j}, GPT-NeoX-20B~\cite{https://doi.org/10.48550/arxiv.2204.06745}, and Falcon-40B~\cite{falcon40b}, all first fine-tuned on the Dolly dataset \cite{DatabricksBlog2023DollyV2}, to study the impact of the size of the model on the performance of CnR. This results in three CnR models: \GPTJDCnR, \GPTNeoXDCnR, and \FalconDCnR. Moreover, we performed CnR fine-tuning on Falcon-40b-Instruct as a strong fine-tuned model to study how much improvement from CnR can be expected when applied to a high-performing LLM. This CnR model is denoted as \FalconICnR.

All the CnR models are trained for 16 epochs with learning rate of $5e-6$ on 4 P4d EC2 instances on Amazon AWS using the Huggingface's transformers library. The value of per device batch size is set to 1, which results in the effective batch size of 32.

\section{Experiments}
\label{sec:exp}


For the evaluation data, we primarily adopt the widely used evaluation dataset of FastChat, which includes 80 queries encompassing diverse topics, such as writing, knowledge acquisition, math, coding, etc.\footnote{https://github.com/lm-sys/FastChat/blob/v0.2.5/fastchat/eval/table/question.jsonl} When evaluating the best CnR setting, we also obtain prompts from Anthropic Helpful~\cite{bai2022training} and Self-Instruct~\cite{wang-etal-2023-self-instruct} as additional evaluation data.

In terms of evaluation metrics, we adopt both automatic evaluation and human evaluation, where each evaluation instance contains a pair of model responses to compare, with three possible outcome labels: win, lose, and tie. For automatic evaluation, we use GPT4 as the evaluator, which has been shown to be able to match the perormance of both controlled and crowd-sourced human preferences~\cite{lin-chen-2023-llm,zheng2023judging}. The detailed prompt for GPT4 evaluation is provided in  Appendix. Considering that GPT4 is known to exhibit a positional bias~\cite{zheng2023judging}, we employ a bidirectional evaluation setting. This means that each evaluation instance with a pair of model responses is evaluated twice by switching the positions of the two responses. For human evaluation, we leverage Amazon Mechanical Turk workers. Each pair of responses is evaluated by 5 randomly assigned human annotators and we only take into account the evaluated pairs where at least 3 out of 5 annotators agree. We then take the majority vote as the final human label. 

For each instance, we obtain the evaluation scores of responses from models A and B. If the score of model A is higher than that of B, then model A gets a win for that evaluation instance. After obtaining the number of wins for model A ($W_A$ ) and for model B ($W_B$), as well as number of ties $T$ on all evaluation instances, the win rate of model A is calculated as $\frac{(W_A+T/2)}{W_A+W_B+T}\times 100$.

\section{Results}

\subsection{Best CnR Setting}


\begin{wraptable}{r}{7.5cm}
\centering
\small
\caption{Win rates (\%) in head-to-head matchups between different formulations, with win rates according to 5 AMT annotators and GPT4.}
\resizebox{0.5\columnwidth}{!}{
\begin{tabularx}{\linewidth}{l l l c c}
\toprule
  \textbf{Comparison} & &  & \textbf{Human} & \textbf{GPT4} \\

 \midrule

  $p \rightarrow i \rightarrow r$ & \textbf{vs} & $p \rightarrow r$ & 58.5 & 68.8 \\
  $p \rightarrow i \rightarrow c \rightarrow r$  & \textbf{vs} & $p \rightarrow r$ & 64.0 & 70.0 \\
  $p \rightarrow i \rightarrow c \rightarrow r$  & \textbf{vs} & $p \rightarrow i  \rightarrow r$ & 54.0 & 56.9 \\
\bottomrule
\end{tabularx}}
\label{tab:cnr_settings}
\end{wraptable}

We first study the best setting, among the three settings presented in Figure \ref{fig:cnr-rec}, to leverage the CnR annotated data for improving model responses.  
The goal is to understand the role of the initial response and critique in CnR.
To this end, we conduct both human and GPT4 evaluation over the revised responses for 486 prompts aggregated from Anthropic Helpful, Self-Instruct, and FastChat. Results of these experiments are shown in Table \ref{tab:cnr_settings}. We can see that the best setting is where the model first generates an initial response, then generates a critique of that response, and then generates a revision of the initial response based on the critique. In general this experiment shows that the improvements from CnR are not simply due to the higher quality of the revised responses, instead enabling the model to generate a critique and revise the initial response contributes to the improvements obtained from CnR.

\subsection{CnR Can Improve ChatGPT}

\begin{figure}[h]
    \centering
    \begin{minipage}{0.47\textwidth}
        \centering
        \includegraphics[width=\textwidth]{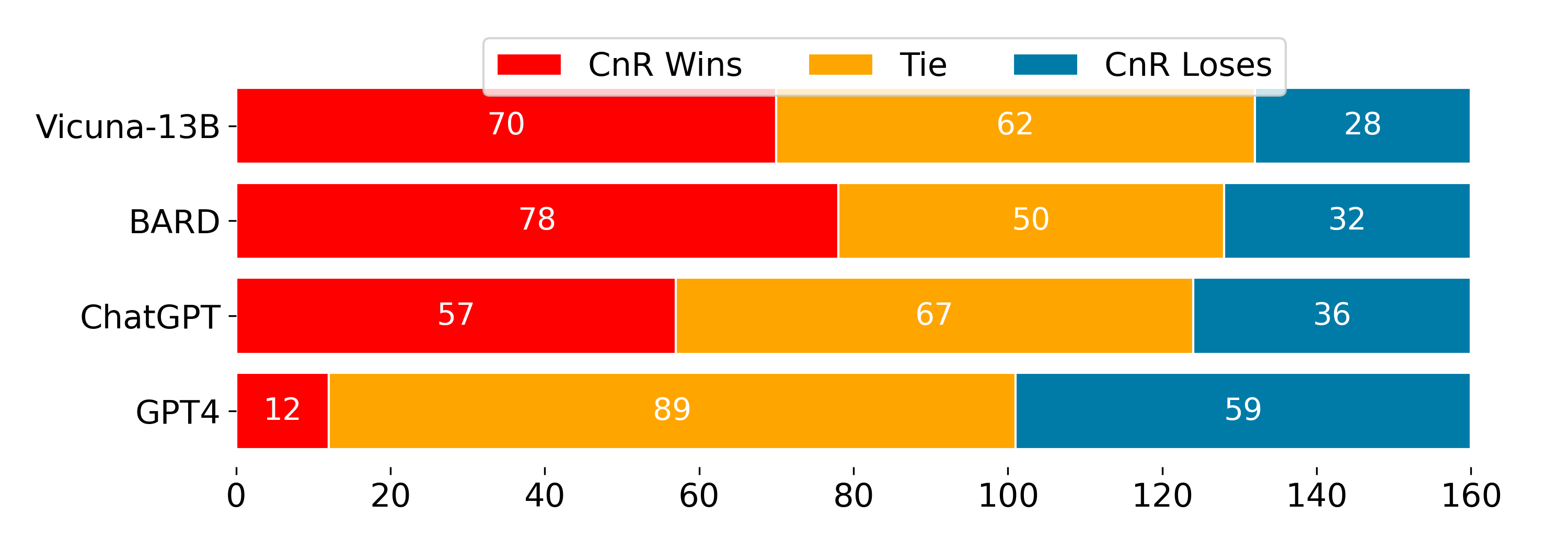}
        \caption{Win rates between the original responses from Vicuna-13B, BARD, ChatGPT, and GPT4 on FastChat 80 queries and the revised responses by \FalconICnR.}
    \label{fig:falcon-40b-I-bar-chart}
    \end{minipage}
    \hfill
    \begin{minipage}{0.47\textwidth}
        \centering
        \includegraphics[width=\textwidth]{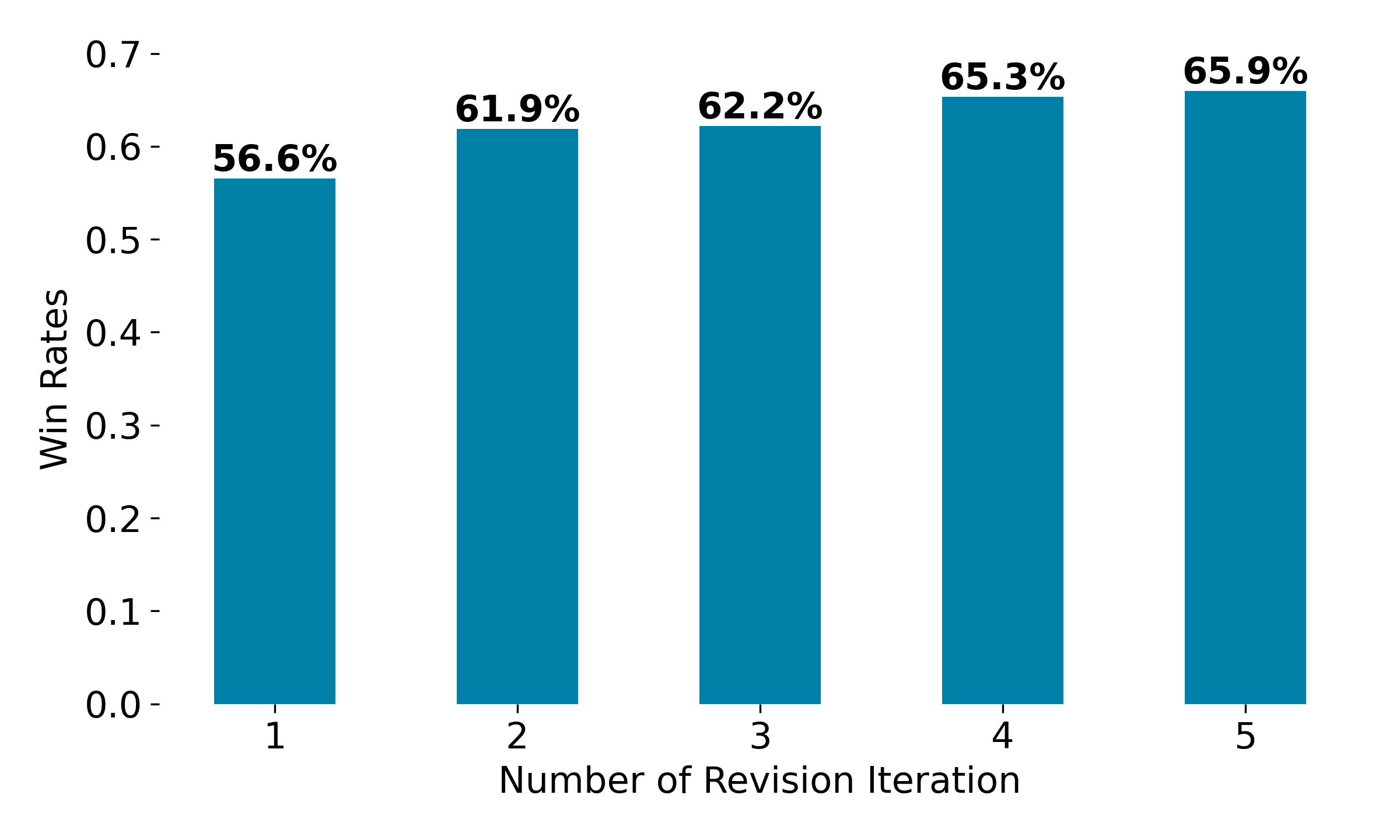}
        \caption{Win rates of revised responses by Falcon-40B-Instruct-CnR with respect to the original responses from ChatGPT over iterative revisions.}
    \label{fig:falcon-40b-instruct-cnr-iterative-revision}
    \end{minipage}
\end{figure}

To show the effectiveness of the CnR method, we use our best CnR model, \FalconICnR (which is fine-tuned on all 1K CnR training samples), to revise responses generated by Vicuna-13B, BARD, ChatGPT, and GPT4 on FastChat test set. Then we use GPT4 as the evaluator to compare the original responses against the revised ones. Win rates are shown in Figure~\ref{fig:falcon-40b-I-bar-chart}, where each comparison pair consists of one original response from a LLM and one corresponding revised response from \FalconICnR. As can be seen, \FalconICnR can effectively improve responses from Vicuna-13B, BARD, and ChatGPT. It should be noted that the base model of \FalconICnR, i.e. Falcon-40B-Instruct, was fine-tuned on data from ChatGPT. This implies that for a model X that has been instruction-tuned on data $\mathcal{D}$, we can further fine-tune it on the CnR data and then use the obtained X-CnR model to further improve the training data $\mathcal{D}$ to a better version $\mathcal{D'}$. Such an improved dataset $\mathcal{D'}$ can potentially help train a better instruction-tuned model X'.

We also conducted human evaluation to verify the advantage of the revised \CHATGPT responses by \FalconICnR over the original ones. We obtain a 58.2\% win rate, which is close to the win rate of 56.6\% from \GPT evaluation.

\subsection{CnR Can Improve Iteratively}

To validate that our CnR method can keep improving responses over iterations, we use \FalconICnR to revise the responses from ChatGPT five times iteratively. In iteration $N$, the revision model revises the responses obtained in iteration $N-1$, or the initial response, if $N=1$. We compare the quality (as measured by GPT4) of the revised responses with the initial ChatGPT responses on the FastChat test set and present the win rates in Figure \ref{fig:falcon-40b-instruct-cnr-iterative-revision}. As can be seen, the win rate has a big jump from iteration 1 to iteration 2 (win rate increases from 56.6\% to 61.9\%), then gradually increases without plateau until iteration 5 (reaching up to 65.9\% win rate). This trend validates that our CnR method can keep improving responses by many iterations, yielding a final improvement of over 10\% in terms of win rate.

\begin{figure*}
    \centering
    \small
    \includegraphics[width=0.9\textwidth]{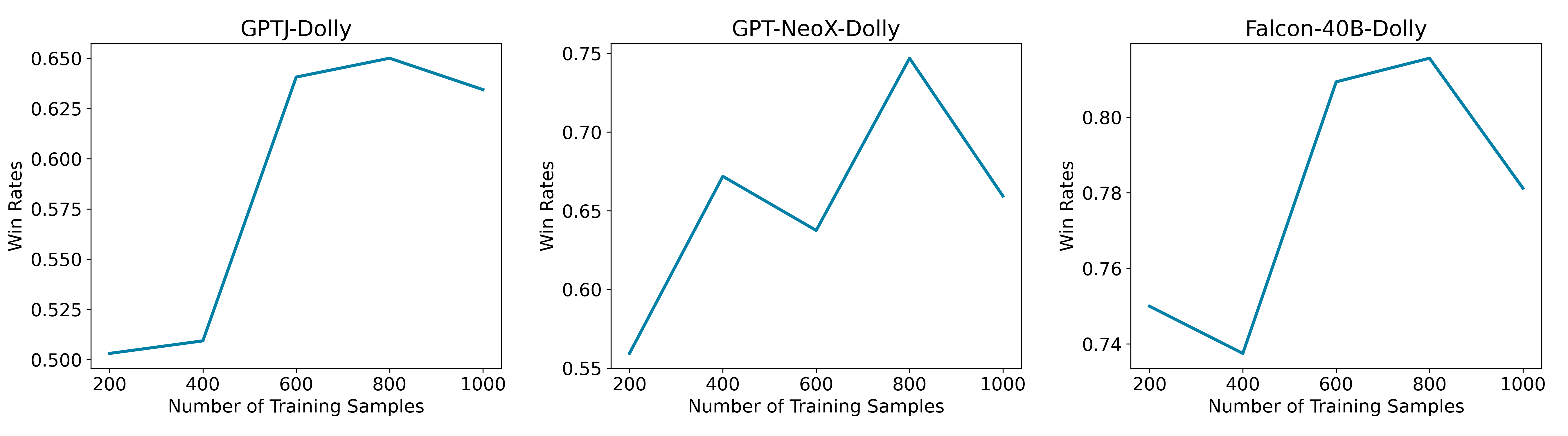}
    \caption{Win rate of the revised responses over the original ones for different base models fine-tuned on different numbers of CnR samples. Original responses are from instruction-tuned model X (such as GPTJ-Dolly in the left-most figure), while revised responses are from the corresponding revision model X-CnR that fine-tunes model X on the CnR data (e.g., GPTJ-Dolly-CnR.)}
    \label{fig:initial_v2_revised_different_models_samples}
\end{figure*}

\subsection{What Impacts Revision Quality?}

We would like to investigate what factors can influence the revision quality of our CnR method. By decomposing the steps of CnR, we have identified the following factors that could play a role: the base pre-trained model size, instruction-tuning quality prior to fine-tuning on CnR samples, and the number of CnR samples.

\paragraph{Instruction Tuning:}
We use both \FalconDCnR and \FalconICnR to revise responses from Vicuna-13B on FastChat test set and calculate the win rates of the revised responses with respect to the original ones. The results are 63.1\% and 53.1\% for \FalconICnR and \FalconDCnR respectively. Both revision models use the same Falcon-40B base model and the same CnR data, and the only difference is the instruction-tuning data. \FalconICnR is based on a set of much larger quantity and higher quality of instruction-tuning data than \FalconDCnR, and it presents much better revision quality, indicating that \textbf{better instruction tuning can help with the revision quality}. 

\paragraph{Number of CnR Samples.}
In order to study the influence of the size of CnR training data, we fine-tune an instruction-tuned model X on different numbers of CnR samples ranging from 200 to 1000 with an interval of 200. We then use the obtained revision model X-CnR to revise the responses from model X and calculate the win rate of the revised responses over the original ones. The results are illustrated in Figure \ref{fig:initial_v2_revised_different_models_samples}. 
Generally, \textbf{the revision quality first improves significantly when sample size increase from 0 to 600, then gradually saturates with 800 samples and over}. 

\paragraph{Base Model Size.}

 As shown in Figure \ref{fig:initial_v2_revised_different_models_samples}, trained on the same number of CnR samples, larger models tend to achieve higher win rate of the revised responses over the original ones. For example, with 800 CnR samples, \FalconDCnR (40B parameters), \GPTNeoXDCnR (20B), and \GPTJDCnR (6B) obtain 81.6\%, 74.7\%, and 65.0\% of win rate, respectively. Such a trend indicates that \textbf{larger pre-trained models lead to better revision quality}.

\section{Discussion}

This section provides additional analysis of the CnR data, which we collected and used to train the CnR models, and the CnR model behavior. 

\subsection{Distribution of Critiques}
\begin{figure*}[ht!]
\centering
\small
\includegraphics[width=1\textwidth]{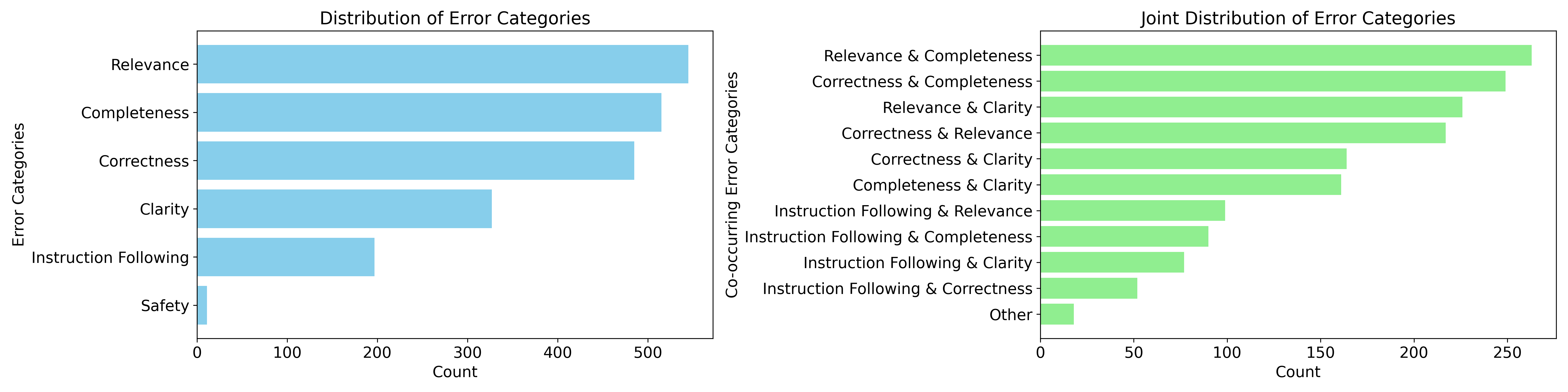}
\caption{Distribution of error categories for the responses mentioned by critiques in the CnR data.}
\label{fig:critique_distribution}
\end{figure*}

We show the distribution of critiques across the 1000 CnR samples.
Specifically, we define six high-level error categories that are mentioned by the critiques (aspects that initial responses are not good at): (i) instruction following, (ii) correctness (factual, logical), (iii) relevance, (iv) completeness, depth and detail, (v) clarity, (vi) safety considerations and limitations. GPT4 is leveraged to annotate whether each human-written critique mentions any of these six error categories. Figure \ref{fig:critique_distribution} shows relevance, completeness, and clarity to be the major error categories. 

\subsection{Critique Quality}

CnR is a multi-step process ($p \rightarrow i \rightarrow c \rightarrow r$), however our prior evaluations consider the prompt and response ($p, r$) and disregard the intermediate critique ($c$). Here we isolate the critique step, $(p, i) \rightarrow c$, in order to evaluate whether CnR models can generate accurate and complete critiques.


To this end, we leverage GPT4 to create a diagnostic dataset for measuring critique quality. Given a particular error category (e.g., instruction following, factual correctness, etc.) and several in-context examples, GPT4 is instructed to generate paired prompts and erroneous responses. The resulting diagnostic set has 551 samples, each consisting of (i) prompt, (ii) response, (iii) error category, (iv) natural language feedback explaining the errors in the response. 

Using this diagnostic test set, we generate critiques with several CnR models. To evaluate the generated critiques, ChatGPT is used to measure the degree (1-5) to which the generated critiques cover the error category and the high-level feedback associated with the prompt and response. 
\begin{table}[!t]
    \centering
    \renewcommand*{\arraystretch}{1.1}
    \small
    \caption{Evaluation of generated critiques on the diagnostic dataset. \CHATGPT was used to evaluate the critique based on the coverage of the ground-truth feedback. We include overall performance, as well as performance on the six error categories: correctness, instruction following, clarity, safety, completeness/detail/depth, relevance.}
    \begin{tabular}{l c c c c c c c}
        \toprule
        \textbf{Name} & \textbf{Overall} & \textbf{Corr.} & \textbf{Instr.} & \textbf{Clr.} & \textbf{Saf.} & \textbf{Compl.} & \textbf{Rel.}\\
        \midrule
        \FalconDCnR & 3.0 & 2.7 & 2.9 & 3.3 & 3.0 & 3.1 & 3.0\\
        \FalconICnR & 3.2 & 2.9 & 3.1 & 3.5 & 3.1 & 3.4 & 3.3 \\
        \CHATGPT & 3.7 & 3.9 & 3.6 & 3.8 & 3.4 & 3.9 & 3.6 \\
        \bottomrule
    \end{tabular}
    \label{tab:critique_analysis}
\end{table}

The results in Table \ref{tab:critique_analysis} show the performance of two CnR models, as well as ChatGPT generated critique. It is noteworthy that our 40B CnR models are shown to perform better at  identifying issues pertaining to clarity and completeness (3.5; 3.4) relative to correctness or instruction following (2.9; 3.1). Since the latter two error categories typically require factual knowledge, this may be a consequence of the base model. Across the six error categories, the difference between \FalconICnR and \FalconDCnR is the largest on completeness (+0.3).

\subsection{Revision Quality}

Similarly, we next aim to isolate and independently evaluate the revision step: $(p, i, c) \rightarrow r$, in order to ascertain whether our CnR models can effectively revise responses according to a provided critique.

\begin{wraptable}{r}{8.1cm}
    \centering
    \renewcommand*{\arraystretch}{1.1}
    \small
    \caption{Evaluation of revision models on the diagnostic dataset. The table shows the adherence of the revisions to (i) detailed critiques, and (ii) low-detail, coarse critiques. The final column shows the adherence of a revision generated based on a coarse critique, to the corresponding detailed critique.}
    \begin{tabular}{l c c c}
        \toprule
        \textbf{Name} & \textbf{Detailed} & \textbf{Coarse} & \textbf{Coarse to Detailed} \\
        \midrule
        \FalconDCnR & 4.86 & 4.85 & 4.77 \\
        \FalconICnR & 4.90 & 4.94 & 4.89 \\
        \CHATGPT & 4.99 & 4.97 & 4.94 \\
        \bottomrule
    \end{tabular}
    \label{tab:revision_analysis}
\end{wraptable}

Starting with the diagnostic dataset used for measuring critique quality, we use ChatGPT to convey the ground-truth high-level feedback as a (i) detailed, actionable critique and (ii) low-detail, coarse critique. The resulting diagnostic dataset for measuring revision quality consists of 551 samples, each containing a (i) prompt, (ii) response, (iii) error category, (iv) coarse critique, (v) detailed critique.

Using this diagnostic test set, we generate revisions with several of our CnR models. To evaluate the generated revisions, ChatGPT is used to measure the degree (1-5) to which the revisions adhere to the critique. Adhering to the critique requires that the revisions resolve the issues raised by the critique, and follow all instructions and specifications mentioned by the critique.

The results shown in Table \ref{tab:revision_analysis} demonstrate that our CnR models effectively adhere to the provided critiques, suggesting that future work should focus more on improving critique generation. The rightmost column of the table assesses whether revisions that were generated based on coarse critiques, adhere to the corresponding detailed critiques. The performance difference between the second column (Detailed) and last column (Coarse to Detailed) indicates the degree to which a particular revision model is robust to under-specified critiques.
This performance drop is smaller for \FalconICnR (-0.01) than for \FalconDCnR (-0.09), suggesting that the continued pre-training of Falcon-40B-Instruct facilitates a more robust revision model.


\subsection{Error Analysis}

\begin{figure}[!h]
    \centering
    \small
    \includegraphics[width=0.8\columnwidth]{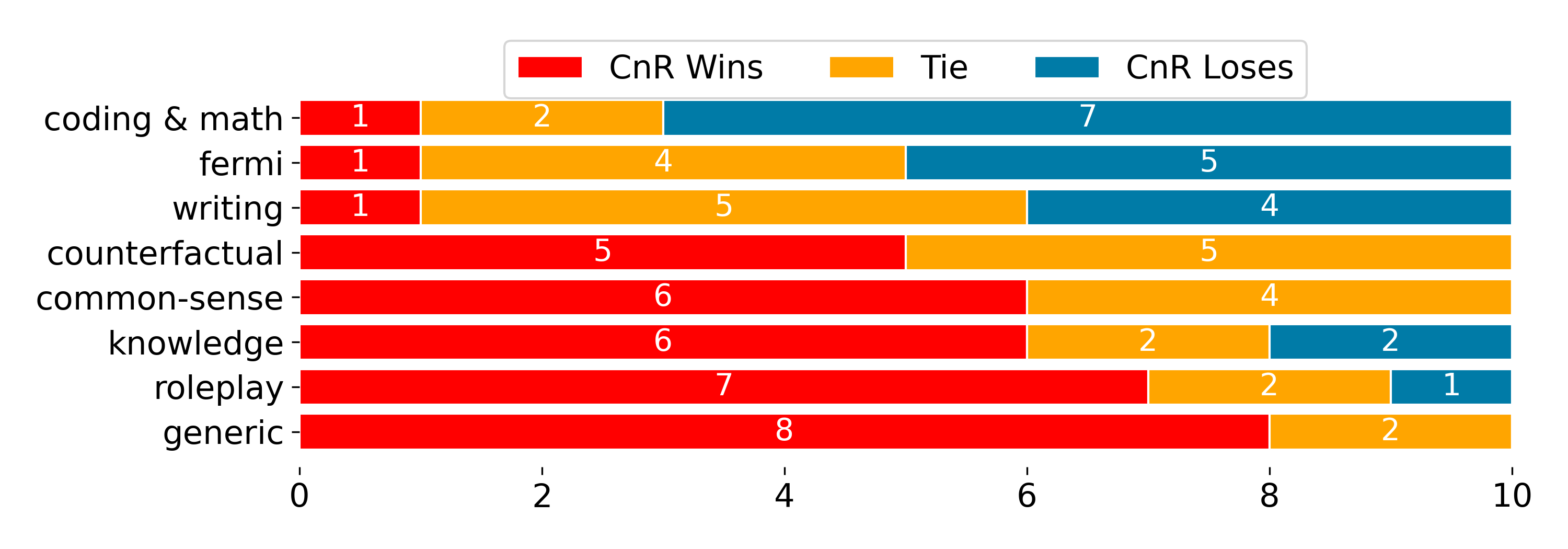}
    \caption{Win rate plot between the original \CHATGPT responses and revised ones by \FalconICnR for each category among 8 on the FastChat test set.}
    \label{fig:falcon-40b-instruct-error-analysis}
\end{figure}

In order to analyze how our CnR model performs on different types of prompts, we categorize the FastChat 80 queries into 8 categories: coding \& math, fermi, writing, counterfactual, common-sense, knowledge acquisition, role play, and generic (each category equally comprises of 10 samples). Figure \ref{fig:falcon-40b-instruct-error-analysis} presents the win rate plot for each category between the original \CHATGPT responses and revised ones by \FalconICnR in  (the average score of two \GPT evaluation scores obtained in both directions is used for comparison). As can be seen, our CnR model can improve responses for the counterfactual, common-sense, knowledge acquisition, role-play, and generic types of prompts, whereas it cannot revise well the coding, math, femi, and writing types of samples. Our CnR data annotation does not include coding, math, and femi types of prompts, which can partly explain the reasons why CnR performs poorly in these types. Besides, all existing open-source LLMs including Falcon-40B (that is used as the base pre-trained model for CnR) still struggle on complex tasks that require advanced reasoning capabilities, e.g. coding, math, physics, etc. 

Regarding the error patterns when our CnR model cannot improve the original responses from \CHATGPT, we find that they are very concentrated on two types of deficits: less detailed, and less structured. Our CnR model may revise a long response from \CHATGPT to be shorter and more concise, which is not preferred by the used \GPT evaluation but may be preferred in some use cases such as chit-chat and voice-based personal assistants. In some cases, the revised responses may lose the well-structured format such as listed bulletins. Such degradation on the response format and structure leads to the lower revision quality on the writing category in Figure \ref{fig:falcon-40b-instruct-error-analysis}. We provide qualitative examples for both types of deficits in Appendix. 

\subsection{Limitation}
\label{sec:limitation}


There are several limitations in our current collected CnR data: 1) The prompt categories need to be expanded to include more hard tasks such as coding, math, reasoning, etc; 2) Initial response quality needs to be enhanced by using more advanced LLMs; 3) The current prompt set contains only single-turn prompts, which needs to be expanded to include multi-turn interactions; 4) Annotated revised response quality should also be improved to be more comprehensive and structured. A notable limitation of our CnR model quality is in the base model used for fine-tuning. Stronger LLMs for CnR training can lead to better critique and revision capability.

\section{Conclusions}

In this work, we present a critique-and-revision (CnR) model that can provide a critique including both positive and negative aspects for a response to a prompt, and further generate a revised response according to the critique. Such a CnR model is achieved by fine-tuning an instruction-tuned LLM on less than 1000 CnR samples. Notably, although the CnR model is derived from an open-source LLM that is much weaker than \CHATGPT, it can improve the \CHATGPT responses and can continue the improvement via iterative revisions, reaching up to 65.9\% of win rate after five iterations. This method can open venues to many interesting applications such as improving training data for supervised fine-tuning, improving reward model for RLHF, and beyond.   

\bibliography{neurips_2023}

\begin{thebibliography}{32}
\providecommand{\natexlab}[1]{#1}
\providecommand{\url}[1]{\texttt{#1}}
\expandafter\ifx\csname urlstyle\endcsname\relax
  \providecommand{\doi}[1]{doi: #1}\else
  \providecommand{\doi}{doi: \begingroup \urlstyle{rm}\Url}\fi

\bibitem[Almazrouei et~al.(2023)Almazrouei, Alobeidli, Alshamsi, Cappelli, Cojocaru, Debbah, Goffinet, Heslow, Launay, Malartic, Noune, Pannier, and Penedo]{falcon40b}
Ebtesam Almazrouei, Hamza Alobeidli, Abdulaziz Alshamsi, Alessandro Cappelli, Ruxandra Cojocaru, Merouane Debbah, Etienne Goffinet, Daniel Heslow, Julien Launay, Quentin Malartic, Badreddine Noune, Baptiste Pannier, and Guilherme Penedo.
\newblock {Falcon-40B}: an open large language model with state-of-the-art performance.
\newblock 2023.

\bibitem[Bai et~al.(2022{\natexlab{a}})Bai, Jones, Ndousse, Askell, Chen, DasSarma, Drain, Fort, Ganguli, Henighan, et~al.]{bai2022training}
Yuntao Bai, Andy Jones, Kamal Ndousse, Amanda Askell, Anna Chen, Nova DasSarma, Dawn Drain, Stanislav Fort, Deep Ganguli, Tom Henighan, et~al.
\newblock Training a helpful and harmless assistant with reinforcement learning from human feedback.
\newblock \emph{arXiv preprint arXiv:2204.05862}, 2022{\natexlab{a}}.

\bibitem[Bai et~al.(2022{\natexlab{b}})Bai, Kadavath, Kundu, Askell, Kernion, Jones, Chen, Goldie, Mirhoseini, McKinnon, et~al.]{bai2022constitutional}
Yuntao Bai, Saurav Kadavath, Sandipan Kundu, Amanda Askell, Jackson Kernion, Andy Jones, Anna Chen, Anna Goldie, Azalia Mirhoseini, Cameron McKinnon, et~al.
\newblock Constitutional ai: Harmlessness from ai feedback.
\newblock \emph{arXiv preprint arXiv:2212.08073}, 2022{\natexlab{b}}.

\bibitem[Black et~al.(2022)Black, Biderman, Hallahan, Anthony, Gao, Golding, He, Leahy, McDonell, Phang, Pieler, Prashanth, Purohit, Reynolds, Tow, Wang, and Weinbach]{https://doi.org/10.48550/arxiv.2204.06745}
Sid Black, Stella Biderman, Eric Hallahan, Quentin Anthony, Leo Gao, Laurence Golding, Horace He, Connor Leahy, Kyle McDonell, Jason Phang, Michael Pieler, USVSN~Sai Prashanth, Shivanshu Purohit, Laria Reynolds, Jonathan Tow, Ben Wang, and Samuel Weinbach.
\newblock Gpt-neox-20b: An open-source autoregressive language model, 2022.
\newblock URL \url{https://arxiv.org/abs/2204.06745}.

\bibitem[Campos and Shern(2022)]{campos2022training}
Jon~Ander Campos and Jun Shern.
\newblock Training language models with language feedback.
\newblock In \emph{ACL Workshop on Learning with Natural Language Supervision. 2022.}, 2022.

\bibitem[Chung et~al.(2022)Chung, Hou, Longpre, Zoph, Tay, Fedus, Li, Wang, Dehghani, Brahma, et~al.]{chung2022scaling}
Hyung~Won Chung, Le~Hou, Shayne Longpre, Barret Zoph, Yi~Tay, William Fedus, Eric Li, Xuezhi Wang, Mostafa Dehghani, Siddhartha Brahma, et~al.
\newblock Scaling instruction-finetuned language models.
\newblock \emph{arXiv preprint arXiv:2210.11416}, 2022.

\bibitem[Conover et~al.(2023)Conover, Hayes, Mathur, Xie, Wan, Shah, Ghodsi, Wendell, Zaharia, and Xin]{DatabricksBlog2023DollyV2}
Mike Conover, Matt Hayes, Ankit Mathur, Jianwei Xie, Jun Wan, Sam Shah, Ali Ghodsi, Patrick Wendell, Matei Zaharia, and Reynold Xin.
\newblock Free dolly: Introducing the world's first truly open instruction-tuned llm, 2023.
\newblock URL \url{https://www.databricks.com/blog/2023/04/12/dolly-first-open-commercially-viable-instruction-tuned-llm}.

\bibitem[Dwivedi-Yu et~al.(2022)Dwivedi-Yu, Schick, Jiang, Lomeli, Lewis, Izacard, Grave, Riedel, and Petroni]{dwivedi2022editeval}
Jane Dwivedi-Yu, Timo Schick, Zhengbao Jiang, Maria Lomeli, Patrick Lewis, Gautier Izacard, Edouard Grave, Sebastian Riedel, and Fabio Petroni.
\newblock Editeval: An instruction-based benchmark for text improvements.
\newblock \emph{arXiv preprint arXiv:2209.13331}, 2022.

\bibitem[FitzGerald et~al.(2022)FitzGerald, Ananthakrishnan, Arkoudas, Bernardi, Bhagia, Delli~Bovi, Cao, Chada, Chauhan, Chen, et~al.]{fitzgerald2022alexa}
Jack FitzGerald, Shankar Ananthakrishnan, Konstantine Arkoudas, Davide Bernardi, Abhishek Bhagia, Claudio Delli~Bovi, Jin Cao, Rakesh Chada, Amit Chauhan, Luoxin Chen, et~al.
\newblock Alexa teacher model: Pretraining and distilling multi-billion-parameter encoders for natural language understanding systems.
\newblock In \emph{Proceedings of the 28th ACM SIGKDD Conference on Knowledge Discovery and Data Mining}, pages 2893--2902, 2022.

\bibitem[Fu et~al.(2023)Fu, Peng, Khot, and Lapata]{fu2023improving}
Yao Fu, Hao Peng, Tushar Khot, and Mirella Lapata.
\newblock Improving language model negotiation with self-play and in-context learning from ai feedback.
\newblock \emph{arXiv preprint arXiv:2305.10142}, 2023.

\bibitem[Glaese et~al.(2022)Glaese, McAleese, Tr{\k{e}}bacz, Aslanides, Firoiu, Ewalds, Rauh, Weidinger, Chadwick, Thacker, et~al.]{glaese2022improving}
Amelia Glaese, Nat McAleese, Maja Tr{\k{e}}bacz, John Aslanides, Vlad Firoiu, Timo Ewalds, Maribeth Rauh, Laura Weidinger, Martin Chadwick, Phoebe Thacker, et~al.
\newblock Improving alignment of dialogue agents via targeted human judgements.
\newblock \emph{arXiv preprint arXiv:2209.14375}, 2022.

\bibitem[Kim et~al.(2023)Kim, Bae, Shin, Kang, Kwak, Yoo, and Seo]{kim2023aligning}
Sungdong Kim, Sanghwan Bae, Jamin Shin, Soyoung Kang, Donghyun Kwak, Kang~Min Yoo, and Minjoon Seo.
\newblock Aligning large language models through synthetic feedback.
\newblock \emph{arXiv preprint arXiv:2305.13735}, 2023.

\bibitem[Lin and Chen(2023)]{lin-chen-2023-llm}
Yen-Ting Lin and Yun-Nung Chen.
\newblock {LLM}-eval: Unified multi-dimensional automatic evaluation for open-domain conversations with large language models.
\newblock In \emph{Proceedings of the 5th Workshop on NLP for Conversational AI (NLP4ConvAI 2023)}, pages 47--58, Toronto, Canada, July 2023. Association for Computational Linguistics.
\newblock URL \url{https://aclanthology.org/2023.nlp4convai-1.5}.

\bibitem[Liu et~al.(2023{\natexlab{a}})Liu, Sferrazza, and Abbeel]{liu2023languages}
Hao Liu, Carmelo Sferrazza, and Pieter Abbeel.
\newblock Chain of hindsight aligns language models with feedback.
\newblock \emph{arXiv preprint arXiv:2302.02676}, 2023{\natexlab{a}}.

\bibitem[Liu et~al.(2023{\natexlab{b}})Liu, Yang, Jia, Zhang, Zhou, Dai, Yang, and Vosoughi]{liu2023training}
Ruibo Liu, Ruixin Yang, Chenyan Jia, Ge~Zhang, Denny Zhou, Andrew~M Dai, Diyi Yang, and Soroush Vosoughi.
\newblock Training socially aligned language models in simulated human society.
\newblock \emph{arXiv preprint arXiv:2305.16960}, 2023{\natexlab{b}}.

\bibitem[Lu et~al.(2022)Lu, Welleck, Hessel, Jiang, Qin, West, Ammanabrolu, and Choi]{lu2022quark}
Ximing Lu, Sean Welleck, Jack Hessel, Liwei Jiang, Lianhui Qin, Peter West, Prithviraj Ammanabrolu, and Yejin Choi.
\newblock Quark: Controllable text generation with reinforced unlearning.
\newblock \emph{Advances in neural information processing systems}, 35:\penalty0 27591--27609, 2022.

\bibitem[Madaan et~al.(2023)Madaan, Tandon, Gupta, Hallinan, Gao, Wiegreffe, Alon, Dziri, Prabhumoye, Yang, et~al.]{madaan2023self}
Aman Madaan, Niket Tandon, Prakhar Gupta, Skyler Hallinan, Luyu Gao, Sarah Wiegreffe, Uri Alon, Nouha Dziri, Shrimai Prabhumoye, Yiming Yang, et~al.
\newblock Self-refine: Iterative refinement with self-feedback.
\newblock \emph{arXiv preprint arXiv:2303.17651}, 2023.

\bibitem[Malmi et~al.(2022)Malmi, Dong, Mallinson, Chuklin, Adamek, Mirylenka, Stahlberg, Krause, Kumar, and Severyn]{malmi2022text}
Eric Malmi, Yue Dong, Jonathan Mallinson, Aleksandr Chuklin, Jakub Adamek, Daniil Mirylenka, Felix Stahlberg, Sebastian Krause, Shankar Kumar, and Aliaksei Severyn.
\newblock Text generation with text-editing models.
\newblock \emph{NAACL 2022}, page~1, 2022.

\bibitem[Ouyang et~al.(2022)Ouyang, Wu, Jiang, Almeida, Wainwright, Mishkin, Zhang, Agarwal, Slama, Ray, Schulman, Hilton, Kelton, Miller, Simens, Askell, Welinder, Christiano, Leike, and Lowe]{NEURIPS2022_b1efde53}
Long Ouyang, Jeffrey Wu, Xu~Jiang, Diogo Almeida, Carroll Wainwright, Pamela Mishkin, Chong Zhang, Sandhini Agarwal, Katarina Slama, Alex Ray, John Schulman, Jacob Hilton, Fraser Kelton, Luke Miller, Maddie Simens, Amanda Askell, Peter Welinder, Paul~F Christiano, Jan Leike, and Ryan Lowe.
\newblock Training language models to follow instructions with human feedback.
\newblock In S.~Koyejo, S.~Mohamed, A.~Agarwal, D.~Belgrave, K.~Cho, and A.~Oh, editors, \emph{Advances in Neural Information Processing Systems}, volume~35, pages 27730--27744. Curran Associates, Inc., 2022.
\newblock URL \url{https://proceedings.neurips.cc/paper_files/paper/2022/file/b1efde53be364a73914f58805a001731-Paper-Conference.pdf}.

\bibitem[Pan et~al.(2023)Pan, Saxon, Xu, Nathani, Wang, and Wang]{pan2023automatically}
Liangming Pan, Michael Saxon, Wenda Xu, Deepak Nathani, Xinyi Wang, and William~Yang Wang.
\newblock Automatically correcting large language models: Surveying the landscape of diverse self-correction strategies.
\newblock \emph{arXiv preprint arXiv:2308.03188}, 2023.

\bibitem[Rafailov et~al.(2023)Rafailov, Sharma, Mitchell, Ermon, Manning, and Finn]{rafailov2023direct}
Rafael Rafailov, Archit Sharma, Eric Mitchell, Stefano Ermon, Christopher~D Manning, and Chelsea Finn.
\newblock Direct preference optimization: Your language model is secretly a reward model.
\newblock \emph{arXiv preprint arXiv:2305.18290}, 2023.

\bibitem[Reid and Neubig(2022)]{reid2022learning}
Machel Reid and Graham Neubig.
\newblock Learning to model editing processes.
\newblock \emph{arXiv preprint arXiv:2205.12374}, 2022.

\bibitem[Saunders et~al.(2022)Saunders, Yeh, Wu, Bills, Ouyang, Ward, and Leike]{saunders2022self}
William Saunders, Catherine Yeh, Jeff Wu, Steven Bills, Long Ouyang, Jonathan Ward, and Jan Leike.
\newblock Self-critiquing models for assisting human evaluators.
\newblock \emph{arXiv preprint arXiv:2206.05802}, 2022.

\bibitem[Schick et~al.(2022)Schick, Jane, Jiang, Petroni, Lewis, Izacard, You, Nalmpantis, Grave, and Riedel]{schick2022peer}
Timo Schick, A~Yu Jane, Zhengbao Jiang, Fabio Petroni, Patrick Lewis, Gautier Izacard, Qingfei You, Christoforos Nalmpantis, Edouard Grave, and Sebastian Riedel.
\newblock Peer: A collaborative language model.
\newblock In \emph{The Eleventh International Conference on Learning Representations}, 2022.

\bibitem[Schulman et~al.(2017)Schulman, Wolski, Dhariwal, Radford, and Klimov]{schulman2017proximal}
John Schulman, Filip Wolski, Prafulla Dhariwal, Alec Radford, and Oleg Klimov.
\newblock Proximal policy optimization algorithms.
\newblock \emph{arXiv preprint arXiv:1707.06347}, 2017.

\bibitem[Song et~al.(2023)Song, Yu, Li, Yu, Huang, Li, and Wang]{song2023preference}
Feifan Song, Bowen Yu, Minghao Li, Haiyang Yu, Fei Huang, Yongbin Li, and Houfeng Wang.
\newblock Preference ranking optimization for human alignment.
\newblock \emph{arXiv preprint arXiv:2306.17492}, 2023.

\bibitem[Vernikos et~al.(2023)Vernikos, Bra{\v{z}}inskas, Adamek, Mallinson, Severyn, and Malmi]{vernikos2023small}
Giorgos Vernikos, Arthur Bra{\v{z}}inskas, Jakub Adamek, Jonathan Mallinson, Aliaksei Severyn, and Eric Malmi.
\newblock Small language models improve giants by rewriting their outputs.
\newblock \emph{arXiv preprint arXiv:2305.13514}, 2023.

\bibitem[Wang and Komatsuzaki(2021)]{gpt-j}
Ben Wang and Aran Komatsuzaki.
\newblock {GPT-J-6B: A 6 Billion Parameter Autoregressive Language Model}.
\newblock \url{https://github.com/kingoflolz/mesh-transformer-jax}, May 2021.

\bibitem[Wang et~al.(2023)Wang, Kordi, Mishra, Liu, Smith, Khashabi, and Hajishirzi]{wang-etal-2023-self-instruct}
Yizhong Wang, Yeganeh Kordi, Swaroop Mishra, Alisa Liu, Noah~A. Smith, Daniel Khashabi, and Hannaneh Hajishirzi.
\newblock Self-instruct: Aligning language models with self-generated instructions.
\newblock In \emph{Proceedings of the 61st Annual Meeting of the Association for Computational Linguistics (Volume 1: Long Papers)}, pages 13484--13508, Toronto, Canada, July 2023. Association for Computational Linguistics.
\newblock \doi{10.18653/v1/2023.acl-long.754}.
\newblock URL \url{https://aclanthology.org/2023.acl-long.754}.

\bibitem[Yuan et~al.(2023)Yuan, Yuan, Tan, Wang, Huang, and Huang]{yuan2023rrhf}
Zheng Yuan, Hongyi Yuan, Chuanqi Tan, Wei Wang, Songfang Huang, and Fei Huang.
\newblock Rrhf: Rank responses to align language models with human feedback without tears.
\newblock \emph{arXiv preprint arXiv:2304.05302}, 2023.

\bibitem[Zhao et~al.(2022)Zhao, Khalman, Joshi, Narayan, Saleh, and Liu]{zhao2022calibrating}
Yao Zhao, Mikhail Khalman, Rishabh Joshi, Shashi Narayan, Mohammad Saleh, and Peter~J Liu.
\newblock Calibrating sequence likelihood improves conditional language generation.
\newblock In \emph{The Eleventh International Conference on Learning Representations}, 2022.

\bibitem[Zheng et~al.(2023)Zheng, Chiang, Sheng, Zhuang, Wu, Zhuang, Lin, Li, Li, Xing, et~al.]{zheng2023judging}
Lianmin Zheng, Wei-Lin Chiang, Ying Sheng, Siyuan Zhuang, Zhanghao Wu, Yonghao Zhuang, Zi~Lin, Zhuohan Li, Dacheng Li, Eric Xing, et~al.
\newblock Judging llm-as-a-judge with mt-bench and chatbot arena.
\newblock \emph{arXiv preprint arXiv:2306.05685}, 2023.

\end{thebibliography}
\bibliographystyle{plainnat}

\clearpage

\appendix


\setcounter{table}{0}
\setcounter{figure}{0}
\renewcommand{\thetable}{A.\arabic{table}}
\renewcommand{\thefigure}{A.\arabic{figure}}

\section{GPT4 Evaluation}

The prompt used for \GPT evaluation for general questions is as below:

\lstset{
    style=mystyle,
    basicstyle=\ttfamily\scriptsize,
    backgroundcolor=\color{white},
    stringstyle=\color{black},
    keywordstyle=\color{black},
    breaklines=false,
    keepspaces=false
}
\begin{lstlisting}[language=Python]
[Question]
(*@{\color{codepurple}{\textsc{<question>}}@*)

[The Start of Assistant 1's Answer]
(*@{\color{codepurple}{\textsc{<response 1>}}@*)
[The End of Assistant 1's Answer]

[The Start of Assistant 2's Answer]
(*@{\color{codepurple}{\textsc{<response 2>}}@*)
[The End of Assistant 2's Answer]

[System]
We would like to request your feedback on the performance of two AI assistants in 
response to the user question displayed above.

Please rate the helpfulness, relevance, accuracy, level of details of their responses.
Each assistant receives an overall score on a scale of 1 to 10, where a higher score 
indicates better overall performance.

Please first output a single line containing only two values indicating the scores for
Assistant 1 and 2, respectively. The two scores are separated by a space. In the 
subsequent line, please provide a comprehensive explanation of your evaluation, avoiding
any potential bias and ensuring that the order in which the responses were presented 
does not affect your judgment.
\end{lstlisting}

The prompt used for \GPT evaluation for coding questions is as below:

\lstset{
    style=mystyle,
    basicstyle=\ttfamily\scriptsize,
    backgroundcolor=\color{white},
    stringstyle=\color{black},
    keywordstyle=\color{black},
    breaklines=false,
    keepspaces=false
}
\begin{lstlisting}[language=Python]
[Question]
(*@{\color{codepurple}{\textsc{<question>}}@*)

[The Start of Assistant 1's Answer]
(*@{\color{codepurple}{\textsc{<response 1>}}@*)
[The End of Assistant 1's Answer]

[The Start of Assistant 2's Answer]
(*@{\color{codepurple}{\textsc{<response 2>}}@*)
[The End of Assistant 2's Answer]

[System]
Your task is to evaluate the coding abilities of the above two assistants. They have
been asked to implement a program to solve a given problem. Please review their code
submissions, paying close attention to their problem-solving approach, code  structure,
readability, and the inclusion of helpful comments.

Please ensure that the assistants' submissions:

1. Correctly implement the given problem statement.
2. Contain accurate and efficient code.
3. Include clear and concise comments that explain the code's logic and functionality.
4. Adhere to proper coding standards and best practices.

Once you have carefully reviewed both submissions, provide detailed feedback on their
strengths and weaknesses, along with any suggestions for improvement. You should first
output a single line containing two scores on the scale of 1-10 (1: no code/no sense; 
10: perfect) for Assistant 1 and 2, respectively. Then give extra comments starting 
from the next line.
\end{lstlisting}

The prompt used for \GPT evaluation for math questions is as below:

\lstset{
    style=mystyle,
    basicstyle=\ttfamily\scriptsize,
    backgroundcolor=\color{white},
    stringstyle=\color{black},
    keywordstyle=\color{black},
    breaklines=false,
    keepspaces=false
}
\begin{lstlisting}[language=Python]
[Question]
(*@{\color{codepurple}{\textsc{<question>}}@*)

[The Start of Assistant 1's Answer]
(*@{\color{codepurple}{\textsc{<response 1>}}@*)
[The End of Assistant 1's Answer]

[The Start of Assistant 2's Answer]
(*@{\color{codepurple}{\textsc{<response 2>}}@*)
[The End of Assistant 2's Answer]

[System]
We would like to request your feedback on the mathematical proficiency of two AI 
assistants regarding the given user question displayed above.

First, please solve the problem independently, without referring to the answers 
provided by Assistant 1 and Assistant 2.

Afterward, please examine the problem-solving process of Assistant 1 and Assistant 2 
step-by-step to ensure their correctness, identifying any incorrect steps if present.
Your evaluation should take into account not only the answer but also the problem-solving
steps. Finally, please output a Python tuple containing two numerical scores for Assistant
1 and Assistant 2, ranging from 1 to 10, respectively. If applicable, explain the reasons 
for any variations in their scores and determine which assistant performed better.
\end{lstlisting}

\section{Critique \& Revision Diagnostic Sets}

Table \ref{diag-critique-analysis-examples} shows examples of the diagnostic dataset used to analyze the critique step of CnR. Table \ref{revision-analysis-examples} shows examples of the diagnostic dataset used to analyze the revision step of CnR.

\begin{table*}[h!]
\centering
\renewcommand*{\arraystretch}{1.0}
\begin{tabular}{|p{2.5cm}|p{10.5cm}|}
\hline
\multicolumn{2}{|c|}{\textbf{Example 1}} \\
\hline
\textbf{Prompt} & What are the differences between a resume and a CV? \\
\hline
\textbf{Response} & A resume is a document that summarizes your work experience, education, skills, and accomplishments. It is typically one to two pages long and is used for job applications in the United States and Canada. 

A CV, or curriculum vitae, is a more detailed document that includes a comprehensive history of your academic credentials, research, publications, and other achievements. It can be several pages long and is used for academic or research jobs or when applying for positions outside of North America. \\
\hline
\textbf{Feedback} & The response does not follow the instructions of the prompt. The prompt asked for differences between a resume and a CV. The response provided definitions of each but did not directly contrast the two. \\
\hline
\textbf{Error Category} & Instruction Following \\
\hline 
\multicolumn{2}{|c|}{\textbf{Example 2}} \\
\hline
\textbf{Prompt} & What are the effects of climate change on agriculture? \\
\hline
\textbf{Response} & Climate change can have several effects on agriculture:

1. Changes in precipitation patterns: Changes in rainfall patterns can lead to droughts or floods, which can damage crops and reduce yields.

2. Increased temperatures: Higher temperatures can increase evaporation rates, leading to drier soils and less water for irrigation.

3. Changes in pest and disease patterns: Warmer temperatures can allow pests and diseases to spread to new areas, affecting crop health and yields.

4. Changes in growing seasons: Changes in temperature and rainfall can affect the length and timing of growing seasons, which can impact crop yields.

5. Increased extreme weather events: Climate change can lead to more extreme weather events, such as storms and heatwaves, which can damage crops and reduce yields. \\
\hline
\textbf{Feedback} & The response includes irrelevant information at the end about the need for a coordinated global response to climate change and reducing our carbon footprints. While these are important issues, they are not specifically related to the effects of climate change on agriculture, which is what the prompt asked for. \\
\hline
\textbf{Error Category} & Relevance \\
\hline
\end{tabular}
\caption{Examples from the diagnostic dataset for critique analysis.}
\label{diag-critique-analysis-examples}
\end{table*}

\begin{table*}[h!]
\centering
\renewcommand*{\arraystretch}{1.0}
\begin{tabular}{|p{2.5cm}|p{10.5cm}|}
\hline
\multicolumn{2}{|c|}{\textbf{Example 1}} \\
\hline
\textbf{Prompt} & How does photosynthesis work in plants? \\
\hline
\textbf{Response} & Photosynthesis is the process by which green plants, algae, and some bacteria convert light energy, usually from the sun, into chemical energy in the form of glucose. This process involves six molecules of carbon dioxide and six molecules of water, which are converted into one molecule of glucose and six molecules of oxygen. \\
\hline
\textbf{Feedback} & The response is correct in describing the general process of photosynthesis, but it incorrectly states that the process involves six molecules of carbon dioxide and six molecules of water. The correct ratio is six molecules of carbon dioxide to twelve molecules of water. \\
\hline
\textbf{Error Category} & Correctness \\
\hline
\textbf{Detailed Critique} & Overall Score: 4/5

Positive: The response accurately describes the general process of photosynthesis and its purpose.

Negative: The response contains a minor error in stating that the process involves six molecules of carbon dioxide and six molecules of water, whereas the correct ratio is six molecules of carbon dioxide to twelve molecules of water. This inaccuracy should be corrected to ensure the response is entirely accurate. \\
\hline
\textbf{Low Detail Critique} & Overall Score: 4/5

Positive: The response accurately describes the general process of photosynthesis.

Negative: There is a minor error in the response. \\
\hline
\multicolumn{2}{|c|}{} \\
\hline
\multicolumn{2}{|c|}{\textbf{Example 2}} \\
\hline
\textbf{Prompt} & What are the main causes of the First World War? \\
\hline
\textbf{Response} & The main causes of the First World War were the assassination of Archduke Franz Ferdinand of Austria, the rise of nationalism in various European countries, and the formation of alliances between different countries. \\
\hline
\textbf{Feedback} & The response requires negative feedback about correctness because it did not include one of the main causes of the First World War, which is the militarism and arms race among the major powers. \\
\hline
\textbf{Error Category} & Correctness \\
\hline
\textbf{Detailed Critique} & Overall Score: 2/5

Positive: The response correctly identifies some causes of the First World War, such as the assassination of Archduke Franz Ferdinand of Austria and the rise of nationalism.

Negative: The response fails to mention one of the main causes of the war, which is the militarism and arms race among the major powers. This omission is significant and should be addressed to provide a more comprehensive and accurate response. \\
\hline
\textbf{Low Detail Critique} & Overall Score: 2/5

Positive: The response mentions some causes of the First World War.

Negative: There is a significant cause missing from the response. \\
\hline
\end{tabular}
\caption{Examples from the diagnostic dataset for revision analysis.}
\label{revision-analysis-examples}
\end{table*}

\section{Error Analysis}

Table \ref{error-analysis-qualitative-example} shows one example where even our best CnR model \FalconICnR may revise the original \CHATGPT response to be less detailed and less structured, which is the main error pattern of our CnR model. 

\begin{table*}[h!]
\begin{tabular}{p{1.5cm}|p{11.5cm}}
\hline
Prompt           & How can governments utilize fiscal and monetary policies to combat economic recessions? \\ \hline
\CHATGPT Response & \begin{tabular}[l]{p{11.3cm}}Governments typically use a combination of fiscal and monetary policies to combat economic recessions.\\ \\ 1. Fiscal policies: \\ \\ Fiscal policies refer to the government's decisions related to taxes and government spending. In a recession, governments can use fiscal policies to boost economic activity by:\\ \\ - Increasing government spending: Governments can increase spending on infrastructure projects, social welfare programs, and other public projects to create jobs and stimulate economic activity.\\ \\ - Cutting taxes: Reducing taxes, particularly for low- and middle-income households, can boost consumer spending and encourage businesses to invest and create new jobs.\\ \\ - Providing subsidies: Governments can provide subsidies to businesses and industries in order to encourage investment and growth. \\ \\ 2. Monetary policies: \\ \\ Monetary policies refer to the management of the money supply and interest rates. In a recession, governments can use monetary policies to stimulate economic activity by:\\ \\ - Lowering interest rates: Lower interest rates can make borrowing cheaper and businesses may be more likely to invest in new projects and hire workers.\\ \\ - Quantitative easing: Central banks can engage in quantitative easing by purchasing government bonds or other assets, which can increase the money supply and stimulate economic activity.\\ \\ Overall, the goal of both fiscal and monetary policies is to inject demand into the economy and get people and businesses spending and investing again in order to help lift the economy out of recession.\end{tabular} \\ \hline
Revised Response & \begin{tabular}[l]{p{11.3cm}}Governments can utilize fiscal and monetary policies to combat economic recessions. Fiscal policies refer to the government's decisions related to taxes and government spending, while monetary policies refer to the management of the money supply and interest rates. In a recession, governments can use fiscal policies to boost economic activity by increasing government spending, cutting taxes, and providing subsidies. They can use monetary policies to stimulate economic activity by lowering interest rates and engaging in quantitative easing. The goal of both fiscal and monetary policies is to inject demand into the economy and get people and businesses spending and investing again in order to help lift the economy out of recession. However, it is important to note that there are differences in effectiveness, trade-offs, time lags, and limitations to these policies. \end{tabular} \\ \hline                                                                     
\end{tabular}
\caption{A qualitative example to show that our best CnR model \FalconICnR may revise the original \CHATGPT response to be less detailed and less structured.}
\label{error-analysis-qualitative-example}
\end{table*}







\end{document}